# Infrared image identification method of substation equipment fault under weak supervision

Anjali Sharma, Priya Banerjee, Nikhil Singh
Department of Electrical Engineering, Savitribai Phule Pune University

**Abstract:** This study presents a weakly supervised method for identifying faults in infrared images of substation equipment. It utilizes the Faster RCNN model for equipment identification, enhancing detection accuracy through modifications to the model's network structure and parameters. The method is exemplified through the analysis of infrared images captured by inspection robots at substations. Performance is validated against manually marked results, demonstrating that the proposed algorithm significantly enhances the accuracy of fault identification across various equipment types.

## 0 Introduction

As a useful tool to detect the defects of substation equipment, infrared imaging makes it possible to visualize the status of equipment and identify problems that may cause temperature anomalies[1]. Using infrared image to detect the heating of equipment has the advantages of early detection, no contact, high reliability, low cost and continuous monitoring, so it has gradually become a common method to identify the defects of substation equipment [2,3].

A large number of infrared images are often produced in the process of substation inspection, and the detection speed and efficiency can be greatly improved by using artificial intelligence technology. Because of the loss of color, texture and other information, infrared image recognition is still a difficult problem [4,5]. Therefore, the application of computer vision technology to improve the recognition accuracy of infrared images in substation equipment fault diagnosis has become a widely studied topic in academic circles. Generally speaking, infrared image recognition can be realized by two different methods: traditional image processing method and deep learning algorithm. Traditional methods usually consist of multiple stages, including feature extraction, target location and image segmentation. The performance of this kind of model is greatly influenced by the extracted features, and its robustness is often poor. For example, reference [6] puts forward an improved method of optimal threshold of Canny operator for power equipment detection. However, this method can't deal with complex background, which has great limitations in practical application. Literature [7] proposes an improved watershed algorithm for image segmentation of power equipment, but this method still requires a fixed image background. Reference [8] proposed a dynamic adaptive optimization algorithm to determine the fuzzy parameters in the target identification model. However, the positioning accuracy of this method for abnormal power equipment is not ideal. On the other hand, the deep learning algorithm shows higher accuracy and generalization ability in the target identification task. This is because they directly use the original image for end-to-end training, thus improving the feature extraction ability and robustness. For example, literature [9] further improves the accuracy of power equipment image recognition by using random forest classifier instead of Softmax classifier commonly used in convolutional neural networks.

Taking the defect detection of substation equipment as an example, it is difficult to collect training samples, especially in complex shooting environment, the sample size that meets the requirements is often small, and it is easily affected by extreme weather conditions. In our research, we introduce a novel algorithm tailored for detecting defects in thermal infrared images of electrical substations. Initially, the method utilizes the Faster RCNN framework to pinpoint substation components. It employs a sophisticated backbone network for high-level feature extraction from the images. The region proposal network is fine-tuned to adapt to the unique shapes of substation equipment, while shared convolutional layers are used for both bounding box determination and classification. Next, we propose a method for extracting image features based on the temperature variance observed between normally functioning and defective parts in substations. This approach involves estimating the temperature probability density using kernel functions. Furthermore, we integrate weakly supervised learning techniques. By leveraging a limited set of labeled samples, we create temperature feature prototype vectors specific to different types of equipment. These prototypes are then refined using unlabeled samples, enhancing the overall accuracy of the model. To demonstrate the effectiveness of our algorithm, we present a case study where it was applied to infrared images captured by inspection robots in the field.

# 1 Substation Equipment Identification Based on Faster RCNN

In order to solve this problem, we design a multi-size anchor region generation strategy, which uses three sizes and five aspect ratios. For slender components in substation, such as bushing, post insulator, bus, etc., the accuracy of model identification is obviously improved after adding anchor area with aspect ratio of 0.25. Different combinations of the above dimensions and aspect ratio will generate 15 suggested areas at each sliding window, covering all the target devices in the infrared image comprehensively, thus improving the detection ability of the devices. Secondly, through the sliding convolution of the input image, the mapping from the suggested region to the low-dimensional features is realized. Taking the VGG-16 model [14] as an example, the generated feature dimension is 512.

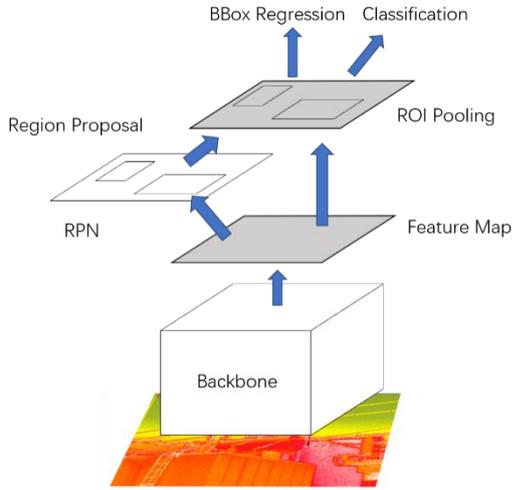

**Fig. 1 Network structure of Faster RCNN**

# 2 Feature extraction based on temperature probability density distribution

Through the Faster RCNN model, we have effectively divided the operating environment and the target equipment, but the accuracy and sensitivity of fault identification directly using the pixel value of the equipment area are not high. Therefore, this paper extracts the temperature probability density distribution of the target equipment area as the bottom feature of fault identification.

The environment of substation can usually be divided into background and equipment. The background usually includes objects such as the sky and brackets that do not generate a lot of heat, while the equipment includes various components. Thermal infrared images have the appearance of ordinary photos, however, each pixel of the former has the temperature information of the corresponding point. After being converted into a three-dimensional image, the output results delineate the temperature distribution. The probability density function of temperature is often used for climate identification [15,16] because it can reflect the temperature characteristics of the investigated object.

$$h(x) = N_x / N \quad (1)$$

$$F(\theta, \theta') = \sum_{x=\theta}^{\theta'} h(x) = \sum_{x=\theta_{min}}^{\theta'} h(x) - \sum_{x=\theta_{min}}^{\theta} h(x) \quad (2)$$

Where: $\theta_{min}$ is the lowest temperature. Obviously, there is one for the whole image $F(\theta_{min}, \theta_{max}) = 1$, where $\theta_{max}$ is the highest temperature.

For the statistically obtained data, it is necessary to estimate the probability density function by using the kernel density estimation method [17.18]. By using the kernel function $K(x)$, the probability that a sample is located in a specific region can be quantified. It should be noted that the choice of kernel function largely determines the performance of the model. Too simple kernel function cannot distinguish different types of waveforms, while too complex kernel function is easily disturbed by noise.

$$K\left(\frac{x - x_i}{w}\right) = \frac{1}{\sqrt{2\pi}} \exp\left(-\frac{1}{2}\left(\frac{x - x_i}{w}\right)^2\right) \quad (3)$$

$$k = \sum_{i=1}^{N} K\left(\frac{x - x_i}{w}\right) \quad (4)$$

$$f(x) = \frac{1}{Nw} \sum_{i=1}^{N} K\left[\frac{x - x_i}{w}\right] \quad (5)$$

This provides a priori understanding for the extraction of temperature distribution in the equipment area, and serves as the classification basis under the condition of weak supervision in the subsequent stage.

# 3 Fault infrared image recognition method under weak supervision

Considering that the devices belonging to the same category show similar temperature distribution, this paper takes the temperature probability distribution as the key feature of clustering and identifying the devices to be detected. The core principle of this algorithm is to determine the initial clustering center by using labeled data, and then classify unlabeled data according to the determined center [19]. The prediction result is determined by calculating the distance to the centers of different categories. In order to solve the problem of low reliability of category center estimation caused by small sample size, this paper uses unlabeled data to adjust category center. In addition, in order to distinguish unlabeled data from labeled data, this paper introduces a confidence parameter $\alpha$, namely

$$c'_m = \alpha c_m + (1-\alpha)\frac{1}{|S'_m|}\sum_{\hat{y}^q_j=m} v^q_j \quad (6)$$

$$c_m = \frac{1}{|S_m|}\sum_{y^s_l=m} v^s_l \quad (7)$$

$$\hat{y}^q_j = \arg\max_m \text{Prob}(y^q_j = m | v^q_j) \quad (8)$$

$$\text{Prob}(y^q_j = m | v^q_j) = \frac{\exp(-d(v^q_j, c_m))}{\sum_{m'}\exp(-d(v^q_j, c_{m'}))} \quad (9)$$

$$d(v^q_j, c_m) = \|v^q_j - c_m\|_2 \quad (10)$$

Where: $v^s_l, v^q_j$ are the characteristic vectors of temperature probability density function corresponding to labeled and unlabeled pictures, $S_m$ is the labeled sample set belonging to category $m$ and $\hat{y}^q_j$ is the predicted result corresponding to unlabeled sample $v^q_j$, where the negative number of Euclidean distance is used as the input calculation probability of Softmax function, $S'_m$ is the unlabeled sample set whose predicted result belongs to category $m$, and $c_m, c'_m$ are the category centers only considering labeled data and the category centers after adding unlabeled data for correction. In this paper, the corrected category center will be used to identify the fault infrared image.

## 4 Case Study

### 4.1 Experimental setup

The data set used in this paper includes 500 infrared images of substation power equipment taken by inspection robot on the spot, and marked manually. The pixel resolution of each picture is $1920\times1080$. The types of power equipment include transformers, bushings, voltage transformers, current transformers and lightning arresters, among which transformers include fans, porcelain bottles and oil conservator. Figure 2 shows some scenarios in the dataset. Among them, the training set under weak supervision includes 150 labeled data and 150 unlabeled data, while only labeled data is used for training under supervised conditions. In addition, the infrared image contains many different faults, such as lightning arrester fault, voltage transformer heating, current transformer heating, bushing joint fault, transformer tap changer fault, transformer lead interface fault and so on. In order to calculate the recognition accuracy of the model, this paper divides different types of equipment into normal and fault types, that is, a total of 10 subcategories.

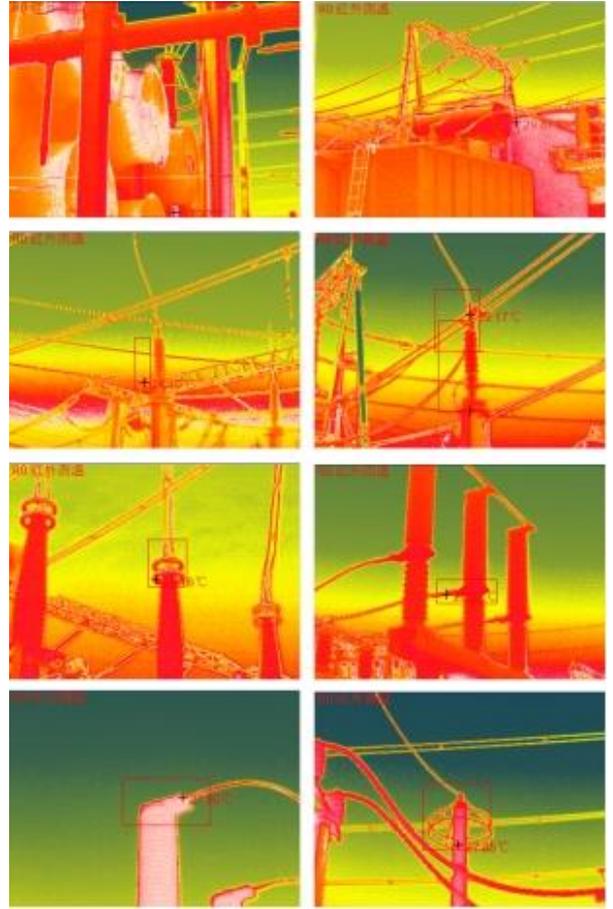

**Fig. 2 Substation infrared image dataset**

### 4.2 Experimental results

The experimental platform is configured as follows: CPU: AMD R95950x, memory: 32G, graphics card: GeForce RTX3060 TI, and software platform: Ubuntu 18.04. The Faster RCNN model adopts pytorch framework, and the basic network adopts ResNet 50 model pre-trained on ImageNet data.

It should be pointed out that infrared image recognition is difficult, and the inspection task mainly focuses on the near target equipment, so this paper mainly considers the near target when training and testing the model. The purpose of device identification is to distinguish the background from the device, which lays the foundation for subsequent feature extraction. If this step is not carried out, the temperature probability density feature is extracted directly, and the feature vector will contain a lot of background information, thus reducing the accuracy of fault identification of the model.

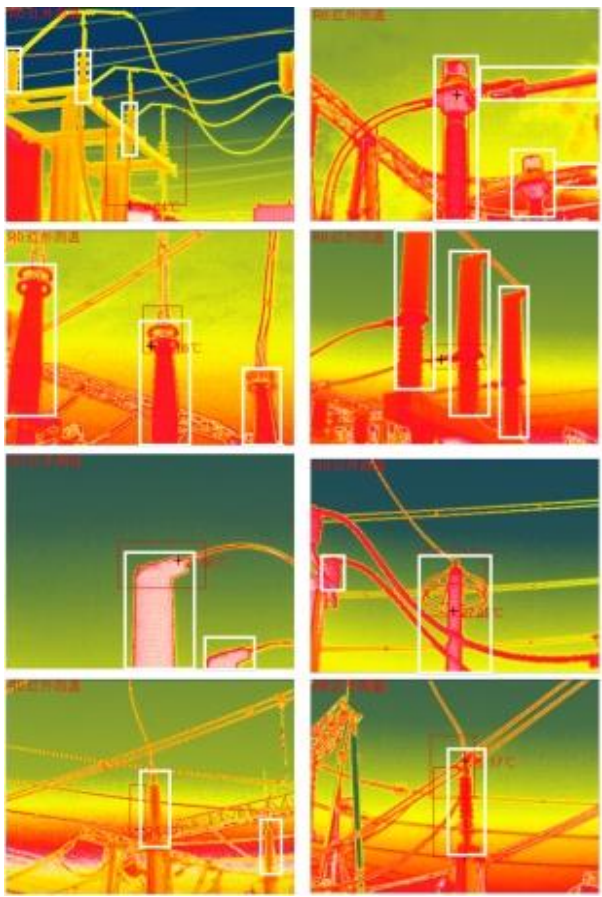

**Fig. 3 Power equipment recognition results based on Faster RCNN**

On the basis of image recognition, we estimate the temperature probability density of the equipment area. We discuss two example cases. The detection object is lightning arrester, the average temperature in the normal area of the equipment is $13.9\,°C$, and the temperature in the fault part is $15.1\,°C$; The detection object is the high-voltage bushing of the main transformer, and the average temperature of the normal area of the equipment is $37.8\,°C$, and the temperature of the fault part is $44.0\,°C$.

Table 1 Recognition accuracy under supervised learning

| Type | device status | | |
|---|---|---|---|
| | normal | breakdown | average |
| transformer | 0.805 | 0.765 | 0.758 |
| casing | 0.901 | 0.815 | 0.839 |
| Voltage transformer | 0.868 | 0.874 | 0.897 |
| Current transformer | 0.851 | 0.865 | 0.860 |
| lightning arrester | 0.874 | 0.812 | 0.873 |
| entirety | 0.837 | 0.832 | 0.844 |

Table 2 Recognition accuracy under weakly-supervised learning

| Type | device status | | |
|---|---|---|---|
| | normal | breakdown | average |
| transformer | 0.869 | 0.859 | 0.854 |
| casing | 0.927 | 0.911 | 0.912 |
| Voltage transformer | 0.936 | 0.923 | 0.937 |
| Current transformer | 0.928 | 0.910 | 0.926 |
| lightning arrester | 0.913 | 0.905 | 0.896 |
| entirety | 0.906 | 0.894 | 0.913 |

From the results of temperature probability density estimation, it can be seen that the temperature distribution in the equipment area is obviously different, and the direct use of the pixel values in the original area will greatly interfere with the subsequent equipment state classification. Using the temperature probability density function as input can effectively distinguish the normal area from the fault part, and at the same time reflect the size difference between them, thus improving the accuracy of model fault identification. In addition, compared with the original function, kernel function estimation can filter the noise in the image, and this method can effectively capture the approximate shape of temperature probability distribution, thus improving the accuracy of subsequent state recognition. By introducing unlabeled data, the robustness of prototype vector calculation can be improved, and then the recognition accuracy can be improved.

# 5 Conclusion

With the continuous improvement of inspection quality requirements for substation equipment, how to realize infrared image defect detection by using artificial intelligence method has become a research hotspot.

In this study, an enhanced version of the Faster RCNN model is utilized for identifying substation equipment, followed by the application of a temperature probability distribution kernel function for more efficient feature extraction. This approach surpasses traditional image processing and deep learning techniques in terms of model performance and data utilization.

For the temperature features extracted from the equipment, a multilayer perceptron is employed to map these features onto prototype vectors that represent the equipment's status, based on labeled samples. These prototype vectors are then refined using unlabeled samples, which improves the model's ability to generalize. The model introduced in this research outperforms existing methods, showing a 6.7% increase in overall average recognition accuracy, and a notable enhancement in the recognition accuracy for each equipment type.

The findings and experimental outcomes of this research clearly demonstrate that the weakly supervised learning algorithm developed here is highly effective in detecting various types of defects in substation equipment through infrared imaging. This has significant implications for the development and upkeep of intelligent substations.